\documentclass[10pt,twocolumn,letterpaper]{article}

\usepackage{cvpr}
\usepackage{times}
\usepackage{epsfig}
\usepackage{graphicx}
\usepackage{amsmath}
\usepackage{amssymb}
\usepackage{array}
\usepackage{multirow, nicefrac}

\usepackage[margin=4pt,font=footnotesize,labelfont=bf,tableposition=top]{caption}
\usepackage[pagebackref=true,breaklinks=true,letterpaper=true,colorlinks,bookmarks=false]{hyperref}

\cvprfinalcopy %

\usepackage{xspace}
\def\OurMethod{HiNAS\xspace}

\ifcvprfinal\fi
\begin{document}

\title{Memory-Efficient Hierarchical Neural Architecture Search for Image Denoising\thanks{This work was done when H. Zhang was visiting The University of Adelaide. Correspondence: C. Shen ($\sf chunhua.shen@adelaide.edu.au$).}
}

\author{
Haokui Zhang$ ^{\dag\ddag} $,
~ Ying Li$^\dag$,
~ Hao Chen$^\ddag$,
~ Chunhua Shen$^\ddag$
\\
$^\dag$ Northwestern Polytechnical University, China
~ ~ ~
$^\ddag$ The University of Adelaide, Australia
}

\maketitle
\thispagestyle{empty}

\begin{abstract}
Recently, neural architecture search (NAS) methods have attracted much attention and outperformed manually designed architectures on a few high-level vision tasks. In this paper, we propose \OurMethod (Hierarchical NAS), an effort towards employing NAS to automatically design effective neural network architectures for image denoising. \OurMethod adopts gradient based search strategies  and employs operations with adaptive receptive field to build an flexible hierarchical search space. During the search stage, \OurMethod shares cells across different feature levels to save memory and employ an early stopping strategy to avoid the collapse
issue in NAS,
and considerably  accelerate the search speed. The proposed \OurMethod is both memory and computation efficient, which takes only about 4.5 hours for searching using a single GPU. We evaluate the effectiveness of our proposed \OurMethod on two different datasets,
namely
an additive white Gaussian noise dataset BSD500, and
a realistic noise dataset SIM1800. Experimental results show that the architecture found by \OurMethod has fewer parameters and enjoys a faster inference speed, while achieving highly competitive performance compared with state-of-the-art methods. We also present analysis on the architectures found by NAS.
\OurMethod also shows good performance on experiments for image de-raining.
\end{abstract}

\section{Introduction}

Single image denoising is an important task in low-level computer vision, which restores a clean image from a noisy one. Owing to the fact that
noise corruption always occurs in the image sensing process and may degrade the visual quality of collected images, image denoising
is needed for
 various
computer vision tasks ~\cite{chatterjee2009denoising}.

Traditional image denoising methods generally focus on modeling natural image priors
and use the priors to restore the clean image, including sparse models ~\cite{dong2012nonlocally, mairal2009non}, Markov random field models ~\cite{lan2006efficient}, etc. One drawback of these methods is that most of them involve a complex optimization problem
and
can be time-consuming for
inference
\cite{dabov2007image, gu2014weighted}. Recently, deep learning models have been successfully applied in various computer vision tasks and set new state-of-the-art. Motivated by this, most
recent works on image denoising have shifted
their approaches to deep learning, which builds
a mapping function  from noisy images to the desired corresponding clean images with deep learning models and have often outperformed conventional methods significantly ~\cite{mao2016image, tai2017memnet, Tobias2018Neural, liu2019dual}. Nonetheless, discovering state-of-the-art neural network architectures requires substantial efforts.

Recently
a growing interest is witnessed in developing algorithmic solutions to automate the manual process of architecture design. Architectures  automatically found by algorithms have achieved highly competitive performance
in high-level vision tasks such as image classification ~\cite{zoph2016neural}, object detection \cite{ghiasi2019fpn, NASFCOS} and semantic segmentation ~\cite{liu2019auto,nekrasov2019fast}. Inspired by this, here  we
design algorithms to  automatically search for
neural architectures \textit{efficiently} for  image denoising tasks.
Our main contributions
are summarized as follows.
\begin{enumerate}
\itemsep -0.125cm

\item Based on gradient based search algorithms, we propose a memory-efficient hierarchical neural architecture search approach for image denoising, termed \OurMethod. To our knowledge, this is the first attempt to apply differentiable architecture search algorithms to low-level vision tasks.

\item The proposed \OurMethod is able to search for both inner cell structures and outer layer widths. It is also memory and computation efficient, taking only about 4.5 hours for searching with a single GPU.

\item We apply our proposed \OurMethod on two denoising datasets of different noise modes for evaluation. Experiments show that the
networks found by our \OurMethod{}  achieves highly competitive performance compared with state-of-the-art algorithms, while having fewer parameters and a faster speed.

\item We conduct comparison experiments to analyse the network architectures
found by our NAS algorithm in terms of the internal structure,
offering some insights in architectures found by NAS.

\end{enumerate}

\subsection{Related Work}
\noindent
{\bf CNNs for image denoising}.
To date,
due to the popularity of convolutional neural networks (CNNs), image denoising algorithms have achieved a significant performance boost. Recent network models such as DnCNN \cite{zhang2017beyond} and IrCNN \cite{zhang2017learning} predict the residue presented in the image instead of the denoised image, showing
promising
performance. Lately, FFDNet \cite{zhang2018ffdnet} attempts to address spatially varying noise by appending noise level maps to the input of DnCNN. NLRN \cite{liu2018non} incorporates non-local operations into a recurrent neural network (RNN) for image restoration.  N3Net \cite{Tobias2018Neural} formulates a differentiable version of nearest neighbor search to further improve DnCNN. DuRN-P ~\cite{liu2019dual} proposes a new style of residual connection, where two residual connections are employed to exploit the potential of paired operations.
Some algorithms focus on denoising for real-noisy images. CBDNet \cite{guo2019toward} uses a simulated camera pipeline to supplement real training data. Similar work in \cite{jaroensri2019generating} proposes a camera simulator that aims to accurately simulate the degradation and noise transformation performed by camera pipelines.

\noindent {\bf Network architecture search (NAS)}.
NAS aims to
design
automated  approaches  for
discovering high-performance
neural architectures
such that the procedure of tedious and heuristic
manual
design of neural architectures can be eliminated
from the deep learning pipeline.
Early attempts employ evolutionary algorithms (EAs) for optimizing  neural architectures and parameters. The best architecture may be  obtained  by iteratively mutating a population of candidate architectures~\cite{liu2017hierarchical}. An alternative to EA is to use reinforcement learning (RL) techniques, e.g., policy gradients~\cite{zoph2018learning,NASFCOS} and Q-learning~\cite{zhong2018practical},
to train a recurrent neural network that acts as a meta-controller to generate
potential architectures---%
typically encoded as sequences---%
by exploring a predefined search space.
However,
EA and RL based methods are inefficient in search,  often requiring a large amount of computations. %
Speed-up techniques %
are therefore proposed to remedy this issue. Exemplar works include
hyper-networks \cite{zhang2018graph}, network morphism  \cite{elsken2018efficient} and shared weights \cite{pham2018efficient}.

In terms of the design of search space and search strategies,
our work is most closely related to DARTS~\cite{liu2018darts}, ProxylessNAS~\cite{cai2018proxylessnas} and Auto-Deeplab~\cite{liu2019auto}. DARTS is based on the continuous relaxation of the architecture representation, allowing efficient search of the cell architecture using gradient descent,
which has achieved competitive performance.
Motivated  by this search efficiency, here
we also use the gradient based approach as our search strategy. In addition, we employ convolution operations with adaptive receptive field in building our search space.
We then extend the search space to include widths for cells by layering multiple candidate paths. Another optimization based NAS approach that has  widths included  in its search space is ProxylessNAS. However, it is limited to discover sequential structures and chooses  kernel widths within manually designed blocks (Inverted Bottlenecks~\cite{he2016identity}). By introducing multiple paths of different widths, the search space of our \OurMethod resembles Auto-Deeplab. The three major differences are:
 1)
to retain high resolution feature maps, we do not downsample the feature maps but reply on automatically selected dilated convolutions and deformable convolutions to adapt
the
receptive field;
 2)
we share the cell across different paths which leads to significant memory efficiency, %
only $ \nicefrac{1}{3} $ of that is needed by
Auto-Deeplab counterparts;
3)
to avoid the performance of the selected network degrading  after a certain number of epochs (collapse problem), we employ a simple but effective early stopping search strategy.
In addition, our \OurMethod is proposed for low-level image restoration tasks, the three methods mentioned above are all proposed for high-level image understanding tasks. DARTS~\cite{liu2018darts} and ProxylessNAS~\cite{cai2018proxylessnas} are proposed for image classification. Auto-Deeplab~\cite{liu2019auto}
finds
architectures for semantic segmentation.

Two more relevant works are E-CAE ~\cite{suganuma2018exploiting} and FALSR~\cite{chu2019fast}. E-CAE~\cite{suganuma2018exploiting} employs
EA
to %
search for an architectures of convolutional autoencoders for image inpainting and denoising. FALSR~\cite{chu2019fast} is proposed for super resolution tasks. FALSR combines RL and EA and design a hybrid controller as its model generator. Both E-CAE and FALSR require a relatively large amount of computations
and takes a large amout of GPU time for searching.
Different  from E-CAE and FALSR, our \OurMethod employs gradient based
strategies  in searching for architectures for low-level image restoration tasks, probably for the first time,
and shares cells across different feature levels in order to save memory.
Our method  only needs about 4.5 GPU hours to find a high-performing architecture on the  BSD500 dataset %
(see Section~\ref{Sec: comparisons with NAS methods}).

\section{Our
Approach}

Following~\cite{liu2018darts,cai2018proxylessnas}, we employ gradient-based architecture search strategies  in our \OurMethod and we search for a computation cell as the basic block then build the final architecture by stacking the
found
block with different widths.
\OurMethod defines a %
flexible hierarchical search space to design architectures for
image denoising. In this section, we first introduce how to search for architectures of  cells using continuous relaxation and adaptive search space.
Then we explain how to determine the widths via multiple candidate paths and cell sharing. %
Last,
we present our search strategy and our the loss functions.

\subsection{Inner Cell Architecture Search}

\begin{figure}
\begin{center}
\includegraphics[width=2.7 in]{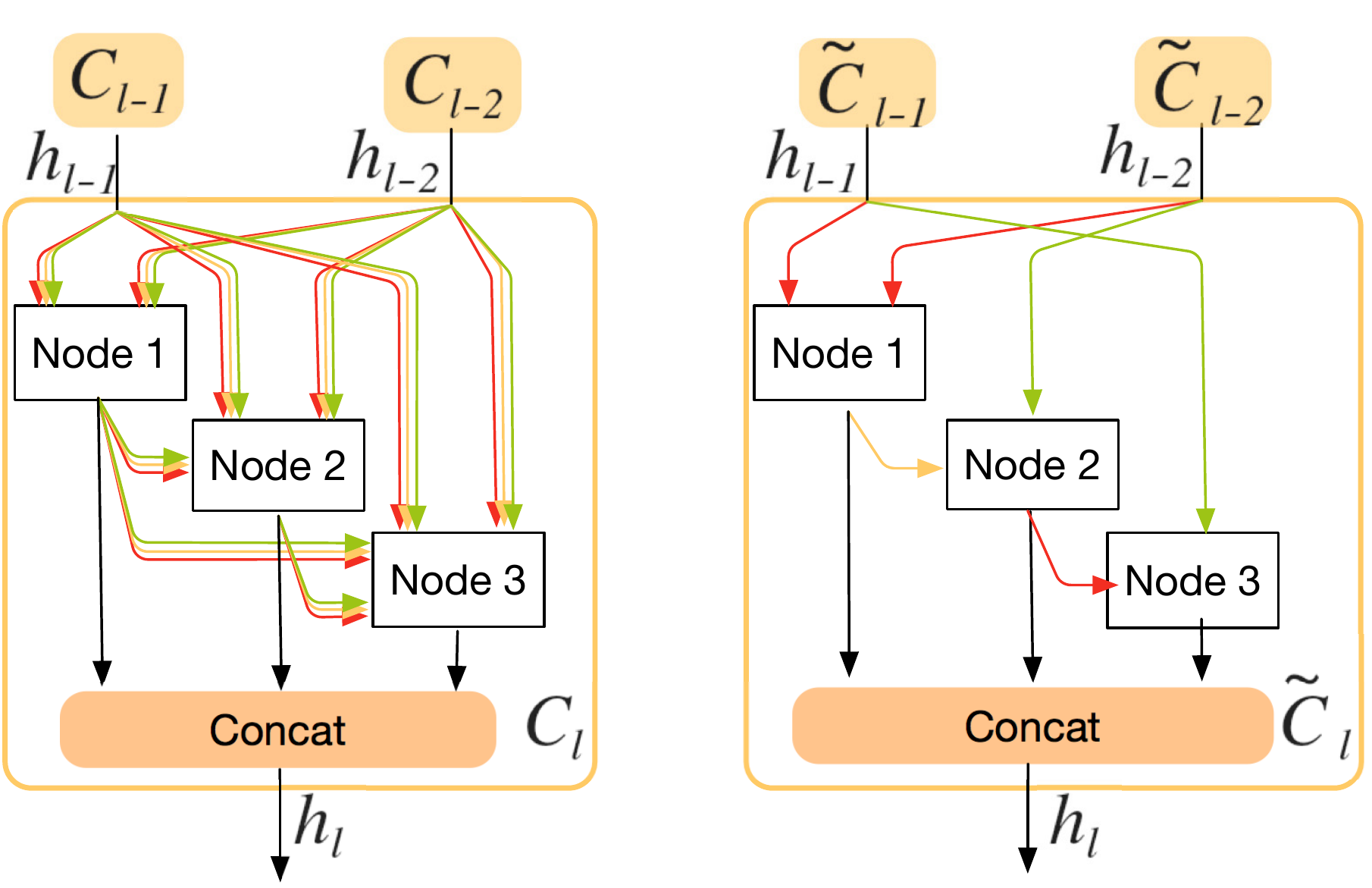}
\end{center}
\vspace{-0.4cm}
\caption{Inner cell architecture search. Left: supercell that contains all possible layer types. Right: the cell architecture search result, a compact cell, where each node only keeps the two most important inputs and each input is connected to the current node with a selected operation.}
\vspace{-0.4cm}
\label{fig:cell_architecture_search}
\end{figure}

\noindent {\bf Continuous relaxation}.
For inner cell architecture search, we employ the continuous relaxation strategy proposed in DARTS~\cite{liu2018darts}. More specifically, we build a supercell that integrates  all possible layer types, which is show in the left side of Figure~\ref{fig:cell_architecture_search}. This supercell is a directed acyclic graph containing a sequence of  $N$ nodes. In Figure~\ref{fig:cell_architecture_search}, we only show three nodes for
clear exposition.

We denote the super cell in layer $l$ as ${C}_{l}$, which takes outputs of previous cells  and the cell before previous cells  as inputs and outputs a  tensor ${h}_{l}$. Inside ${C}_{l}$, each node takes the two inputs of the current cell and the outputs of all previous nodes as input and outputs a  tensor. Taking the $i$th node in ${C}_{l}$ as an example, the output of this node is calculated as follows:
\begin{equation}
{x}_{l,i}=\sum_{{x}_{j}\in{I}_{l,i}}^{}{{O}_{j\rightarrow i }({x}_{j})},
\end{equation}
where ${I}_{l,i}=\{{h}_{l-1}, {h}_{l-2}, {x}_{l,j<i}\}$ is the input set of node $i$. ${h}_{l-1}$ and ${h}_{l-2}$ are the outputs of cells in layers $l-1$ and $l-2$, respectively. ${O}_{j\rightarrow i}$ is the set of possible layer types. Here, to make the search space continuous, we operate each ${O}_{j\rightarrow i}$ in an continuous relaxation fashion, which is:
\begin{equation}
{ O }_{ j\rightarrow i }({ x }_{ j })=\sum _{ k=1 }^{ S }{ { { \alpha  }_{ j\rightarrow i }^{ k } }{ O }^{ k } } ({ x }_{ j }),
\end{equation}
where $\{{O}^{1}, {O}^{2}, \cdots, {O}^{S}\}$ correspond to $S$ possible layer types. ${\alpha}_{j\rightarrow i}^{k}$ denotes the weight of operator ${O}^{k}$.

\vspace{2 pt}
\noindent {\bf Adaptive search space}.
Following several recent image restoration networks~\cite{liu2018non,plotz2018neural,li2018recurrent}, we do not reduce the spatial resolution of the input. To preserve pixel-level information for low-level image processing, we do not downsample the features but rely on operations with adaptive receptive field such as dilated convolutions and deformable convolutions. In this paper, we
pre-define
the following 6 types of basic operators:
\begin{itemize}
\itemsep -0.1cm
    \item conv: $3\times 3$ convolution;
    \item sep: $3\times 3$ separable convolution;
    \item dil: $3\times 3$ convolution with dilation rate of 2;
    \item def: $3\times 3$ deformable convolution v2~\cite{zhu2019deformable};
    \item skip: skip connection;
    \item none: no connection and return zero.
\end{itemize}
Each convolution operation starts with a ReLU activation layer and is followed by a batch normalization layer.

${h}_{l}$ is the concatenation of the outputs of $N$ nodes and it can be expressed as:
\begin{equation}
\begin{split}
{h}_{l}={}& {\rm Cell} ({h}_{l-1}, {h}_{l-2})\\
={}& {\rm Concat} \{{x}_{l,i}|i\in\{1, 2, \cdots, N\}\}.
\end{split}
\end{equation}
In summary, the task of cell architecture search is to learn continuous weights $\alpha$, which
are
updated via gradient descent. After the supercell is trained, for each node, we rank the corresponding inputs according to $\alpha$ values, then
keep
the top two inputs and remove the rest to obtain the compact cell, as shown in the right-side  of Figure~\ref{fig:cell_architecture_search}.

\subsection{Memory-Efficient Width Search}

\begin{figure}[t]
\begin{center}
\includegraphics[width=2.7 in]{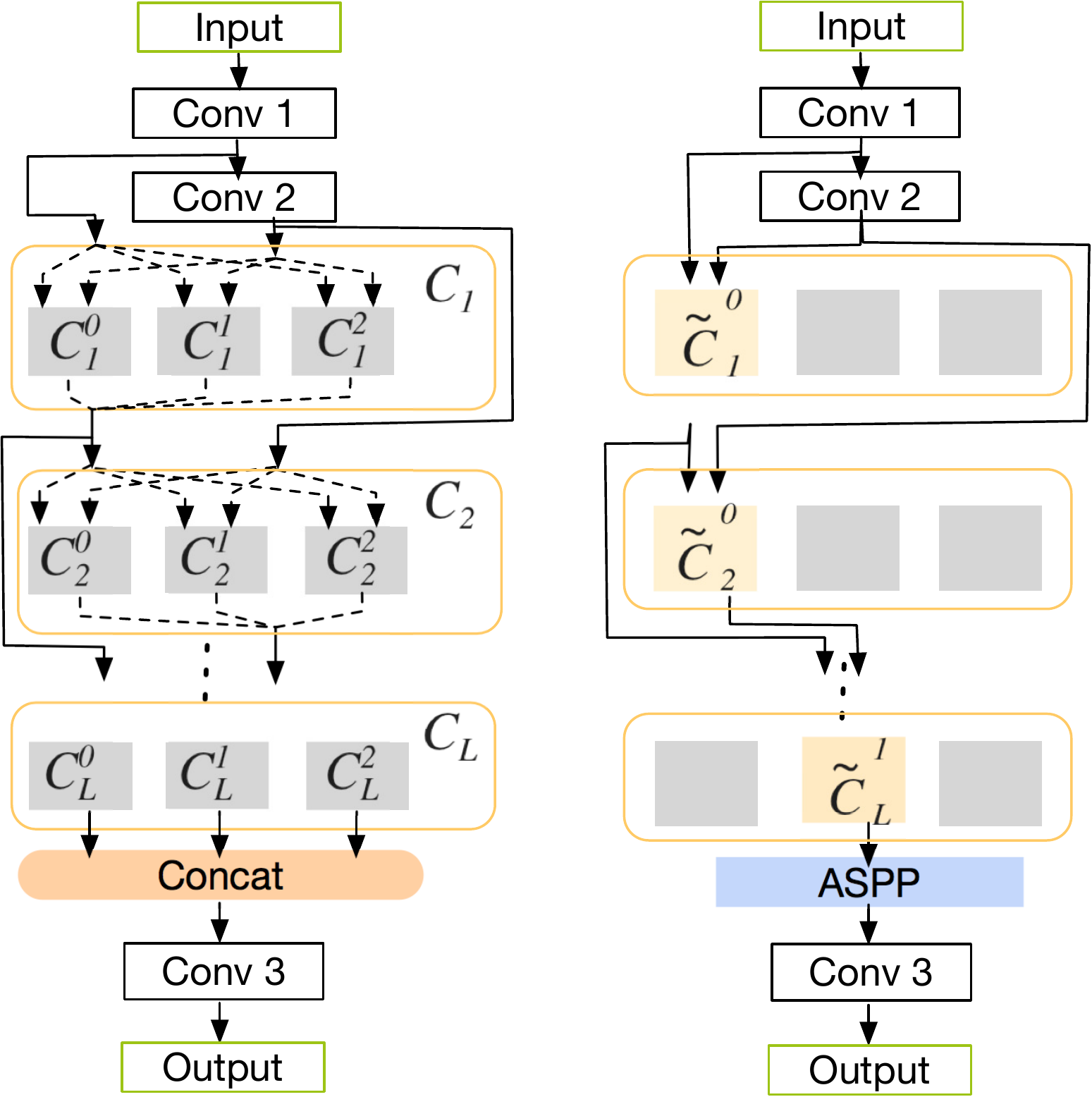}
\end{center}
\vspace{-0.4cm}
\caption{Outer layer width search. Left: network architecture search space, a supernet that consists of supercells and contains several supercells with different widths in each layer. Right: the final architecture obtained from the supernet, a compact network that consists of compact cells and only %
keeps one cell in each layer.}
\vspace{-0.4cm}
\label{fig:network_architecture_search}
\end{figure}

\noindent
{\bf Multiple candidate paths}.
Now we have presented the main idea of cell architecture search, which is used to design the specific architectures inside cells.
As previously mentioned,
the overall network is built by stacking several cells
of
different widths. To build the overall network, we still need to either \textit{heuristically}  set the width of each cell %
or search for a proper width for each cell \textit{automatically}. In conventional CNNs, the change of widths of convolution layers is %
often
related to the change of spatial resolutions.  For instance, doubling the widths of following convolution layers after the features are downsampled. In our \OurMethod, instead of using downsample layers, we rely on operations with adaptive receptive field such as dilated convolutions and deformable convolutions to adjust the receptive field automatically.
Thus the conventional experience of adjusting width no longer applies
to our case.

To solve this problem, we employ the flexible hierarchical search space and leave the task of deciding width of each cell to the NAS algorithm itself, making the search space more general. In fact, several NAS algorithms in the literature also search for  the outer layer width, mostly for high-level image understanding tasks.
For example, FBNet~\cite{wu2019fbnet} and MNASNet~\cite{tan2019mnasnet}%
consider different expansion rates inside their modules to discover compact networks
for image classification.

In this section, we introduce the outer layer width search space which determines the widths of cells in different layers. Similarly, we build a supernet that contains several supercells with different widths in each layer. As illustrated in the left-side of Figure~\ref{fig:network_architecture_search}, the supernet mainly consists of three parts:

1) \textit{start part}, consisting  of input layer and two convolution layer;

2) \textit{middle part}, containing  $L$ layers and each layer having three supercells of different widths;

3) \textit{end part}, concatenating the outputs of ${C}_{L}$, then feeding them to a convolution layer to generate the  output.

Our supernet provides three paths of cells with different widths. For each layer, the supernet decides to increase the width by twice, keeping previous width or reducing the width by two. After searching, only one cell at each layer is kept. The continuous relaxation strategy mentioned in  the cell architecture search section is reused for inter cell search.

At each layer $l$, there are three cells ${C}_{l}^{0}$, ${C}_{l}^{1}$ and ${C}_{l}^{2}$ with widths $W$, $2W$ and $4W$, where $W$ is the basic width and is set to 10 during search phase. The output feature of each layer is
\begin{equation}
{ h }_{ l}=\{{h}_{l}^{0}, {h}_{l}^{1}, {h}_{l}^{2}\},
\end{equation}
where ${h}_{l}^i$ is the output of ${C}_{l}^i$. The channel width of ${h}_{l}^{i}$ is ${2}^{i}NW$, where $N$ is the number of nodes in the cells.

\vspace{2 pt}
\noindent {\bf Cell sharing}.
Each cell $C_l^i$ is connected to $C_{l-1}^{i-1}$, $C_{l-1}^{i}$ and $C_{l-1}^{i+1}$ in the  previous layer and $C_{l-2}^{i}$ two layers before. We first process the outputs $h_{l-1}$ from those layers with a $1\times1$ convolution to form features $f_{l-1}$ with width $2^iW$,  matching the input of $C_l^i$. Then the output for the $i$th cell in layer $l$ is computed with
\begin{equation}
{h}_{l}^{i}=C_l^i \left(
\sum_{k=i-1}^{i+1}{\beta}_{k}^{i}{f}_{l-1}^{k},{f}_{l-2}^{i}
\right),
\label{formula: cell sharing}
\end{equation}
where ${\beta}_{k}^{i}$ is the weight of ${f}_{l-1}^{k}$. We combine the three outputs of ${C}_{l-1}$ according to corresponding weights then feed them to ${C}_{l}$ as input. Here, features ${f}_{l-1}^{i-1}$, ${f}_{l-1}^{i}$ and ${f}_{l-1}^{i+1}$
come from different levels, but they share the cell ${C}_{l}^{i}$ during computing ${h}_{l}^{i}$.

Note the similarity of this design with Auto-Deeplab, which is used to select feature strides for image segmentation. However, in Auto-Deeplab, the outputs from the three different levels are first processed by separate cells with different sets of weights before summing into the output:
\begin{equation}
{h}_{l}^{i}=\sum_{k=i-1}^{i+1}{\beta}_{k}^{i}C_l^{ik}({f}_{l-1}^{k},{f}_{l-2}^{i}),
\label{formula: not share cell}
\end{equation}

\begin{figure}[t]
\begin{center}
\includegraphics[width=2.9in]{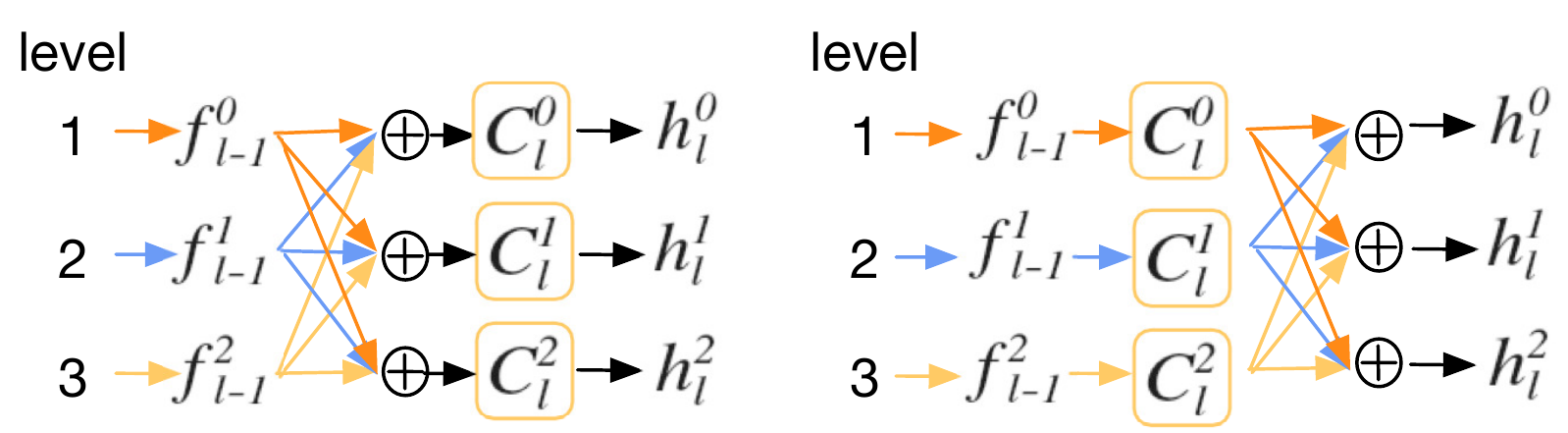}
\end{center}
\vspace{-0.4cm}
\caption{Comparison of cases of whether using cell sharing or not. Left: features from different levels share same cell. Using cell sharing; Right: features from different levels use different cells.}
\vspace{-0.5cm}
\label{fig:cell sharing}
\end{figure}

A comparison between Eqs.~\eqref{formula: cell sharing} and \eqref{formula: not share cell} is shown in Figure~\ref{fig:cell sharing}, where the inputs from layer $l-1$ are not shown out for simplicity.

For the hierarchical structure which has three candidate paths, the cell in each candidate path is used once with Eq.~\eqref{formula: cell sharing} and it is used three times with Eq.~\eqref{formula: not share cell}. By sharing the cell $C_l^i$, we are able to save the memory consumption by a factor of 3  in the supernet,
thus making it possible to use a deeper and wider supernet for more accurate approximations.
This also enables us to use larger
batch sizes during search, accelerating  the search process.

\vspace{2 pt}
\noindent {\bf Deriving final architecture}.
Note that, different from the cell architecture search, we can not simply rank cells of different widths according to $\beta$ values then
keep
the top one cell. In cell widths search, the channel widths of outputs of different cells in the same layer can be very different. Using the strategy that  we have adopted in cell architecture search may lead to the widths of adjacent layers in the final network change drastically, which has a negative
impact on
the efficiency, as explained in~\cite{ma2018shufflenet}. In cell width search, we view the $\beta$ values as probability, then use the Viterbi  decoding algorithm to select the path with the maximum probability as the final result.

\subsection{Searching Using Gradient Descent}

\noindent{\bf Optimization function}.
In terms of the optimization method, our proposed \OurMethod belongs to differentiable architecture search. The searching process is the optimization process. For image denoising, the two most widely used evaluation metrics are PSNR and SSIM~\cite{wang2004image}; and
we design the following loss for optimizing supernet:
\def\ssim{{\rm ssim}}
\def\DartsDn{ {\rm IRNS } }
\begin{equation}
{\rm loss}  = {\left\| {\rm {f}_{net}}(x)-y\right\|}_{2}^{2} + \lambda \cdot  {\rm {l}_{ssim}}({\rm {f}_{net}}(x), y),
\end{equation}{}
where
\begin{equation}
{\rm {l}_{ssim}}(x,y)={\log}_{10}({{\rm ssim}(x,y)}^{-1}),
\end{equation}
Here $x$ and $y$ denote the input image and corresponding ground-truth. ${\rm {l}_{ssim}}(\cdot)$ is a loss item that is  designed to enforce the visible structure of the result. ${\rm {f}_{net}}(\cdot)$ is the supernet. ${\rm ssim}(\cdot)$ is structural similarity~\cite{wang2004image}. $\lambda$ is a weighting coefficient and it is empirically set to 0.5 in all of our experiments.

\vspace{2 pt}
\noindent {\bf Early stopping search}.
During optimizing the supernet with gradient descent, we find that the performance of network founded by \OurMethod is often observed to \textit{collapse} when the number of search epochs becomes large. The very recent method of Darts+~\cite{liang2019darts+}, which is concurrent to this work here, presents similar observations.
Because of this collapse issue,  it is hard to pre-set the  number of search epochs. To solve this problem, we employ an early stopping search strategy. Specifically, we split the training set into three disjoint parts: Train W, Train A and Validation V. Sub-datasets W and A are used to optimize the weights of the supernet (kernels in convolution layers)
and weights of different layer types and cells of different widths ($\alpha$ and $\beta$).  During optimizing, we periodically evaluate the performance of the trained supernet on the validation dataset V. We stop the search procedure when the performance of supernet decreases for a pre-determined number of evaluations.
Then we choose the supernet which offers the highest PSNR and SSIM scores on validation dataset V as the result of the architecture search. Details are
presented
in the search settings of Section~\ref{Sec: implementation details}.

\section{Experiments}

\subsection{Datasets and Implementation Details}
\label{Sec: implementation details}

\noindent \textbf{Datasets}
We carry out the deboising experiments on two datasets.
The first one is BSD500~\cite{martin2001database}. Following~\cite{mao2016image,tai2017memnet,liu2018non, liu2019dual}, we use as the training set the combination of 200 images from the training set and 100 images from the validation set, and test on 200 images from the test set. On this dataset, we generate noisy images by adding white Gaussian noises to clean images with $\sigma=30, 50, 70$.

The second one is SIM1800, built  by ourselves. As the additive white noise models is not able to accurately reproduce the true noise in real world, by using the camera pipeline simulation method proposed in~\cite{jaroensri2019generating}, we build this new denoising dataset SIM1800, which contains 1600 training samples and 212 test samples. Firstly, we use the camera pipeline simulation method to add noises to 25k patches extracted from the MIT-Adobe5k dataset~\cite{bychkovsky2011learning}. We then manually pick
1812 patches which have the most realistic visual effects and finally randomly select 1600 patches as the training set and use the rest as the test set.

\noindent
\textbf{Search settings}. The supernet that we build for image denoising consists of 4 cells and each cell has 5 nodes. we perform architecture search on BSD500 and apply the networks found by \OurMethod on both denoising datasets. Specifically, we randomly choose 2\% of training samples as the validation set (Validation V). The rest are equally divided into two parts:  one part is used to update the kernels of convolution layers (Train W) and the other part is used to optimize the parameters of the neural architecture (Train A).

We train the supernet at most 100 epochs with batch size of 12. We optimize the parameters of kernels and architecture with two optimizers. For learning the kernels of convolution layers, we employ the standard SGD optimizer. The momentum and weight decay are set to 0.9 and 0.0003, respectively. The learning rate decays from 0.025 to 0.001 with the cosine annealing strategy~\cite{loshchilov2016sgdr}. For learning the parameters of an architecture, we use the Adam optimizer, where both learning rate and weight decay are set to 0.001. In the first 20 epochs, we only update the parameters of kernels, then we start to alternately optimize the kernels of convolution layers and architecture parameters from epoch 21.

During the training process of searching, we randomly crop patches of $64\times64$ and feed them to the network. During evaluation, we split each image to some adjacent patches of $64\times64$ and then feed them to the network and finally join the corresponding patch results to otain final results of the whole test image.
We evaluate the supernet for every epoch.

\textbf{Training settings} We train the network for 600k iterations with the Adam optimizer, where the initial learning rate, batchsize are set to 0.05 and 12, respectively. For data augmentation, we use random crop, random rotations $\in \{{0}^{\circ}, {90}^{\circ}, {180}^{\circ}, {270}^{\circ}\}$, horizontal and vertical flipping. For random crop, the patches of $64\times 64$ are randomly cropped from input images.

\subsection{Benefits of Searching for the Outer Layer Width}

\begin{figure}[t]
\begin{center}
\includegraphics[width=2.64 in]{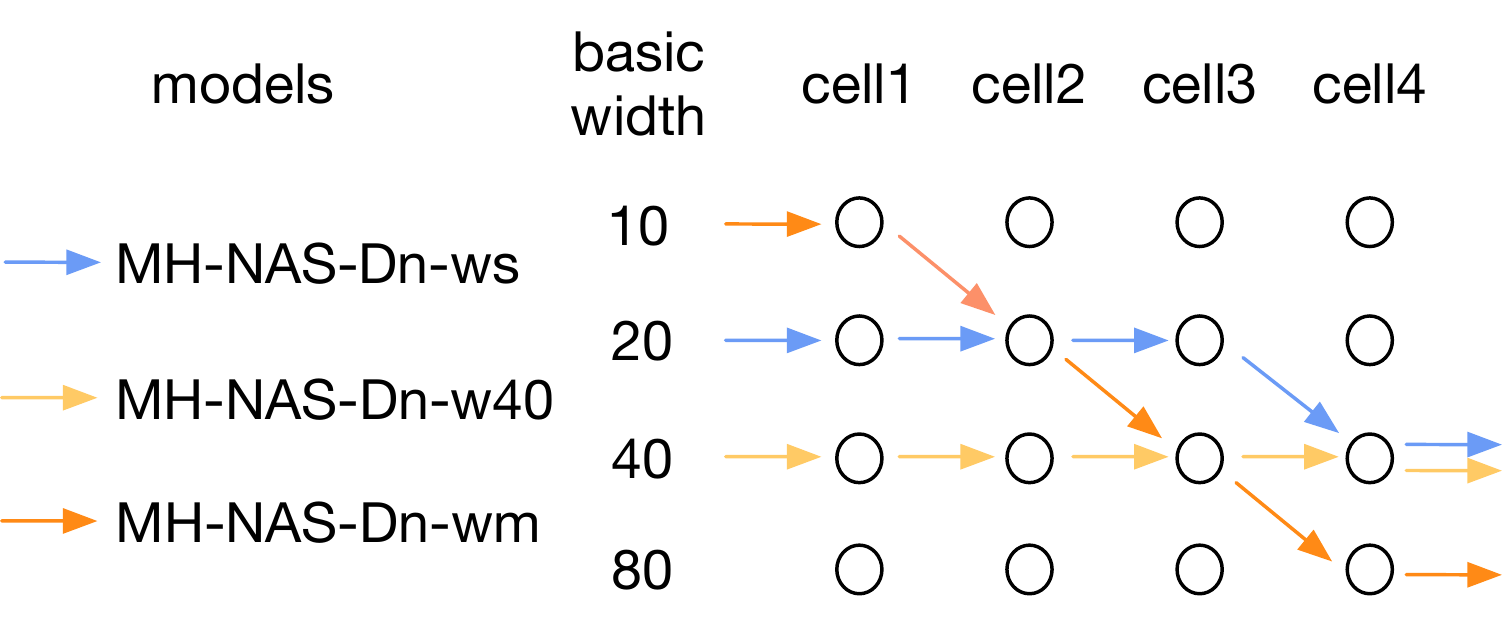}
\end{center}
\vspace{-0.3cm}
\caption{Comparisons of different search settings.}
\label{fig:comparison_width_search}
\end{figure}

\begin{table}[t]
\footnotesize
\renewcommand\arraystretch{1.0}
\begin{center}
{
\begin{tabular}{ c | ccc}
\hline
Models  &  \# parameters (M) & PSNR  & SSIM \\
\hline
\OurMethod-ws      & 0.63  & 29.14  & 0.8403 \\
\OurMethod-w40     & 0.96  & 29.15  & 0.8406 \\
\OurMethod-wm      & 1.13  & 28.89  & 0.8370 \\
\hline
\end{tabular}
}
\end{center}
\vspace{-0.2cm}
\caption{Comparisons of different search settings.}
\vspace{-0.6 cm}
\label{Table: comparisons of different search settings}
\end{table}

In this section, to evaluate the benefits of searching outer layer width, we apply our \OurMethod on BSD500 with three different search settings, which are denoted as \OurMethod-ws, \OurMethod-w40, \OurMethod-wm. For \OurMethod-ws, both inner cell architectures  and out layer width are found by our \OurMethod algorithm. For the latter two settings, only the inner cell architectures are found by our algorithm and the outer layer widths are set manually. The basic width of each cell are set to 40 for \OurMethod-w40. In \OurMethod-wm, we set the basic width of the first cell to 10, then double the basic width cell by cell. The three settings are shown in Figure~\ref{fig:comparison_width_search}. The comparison results for denoising on BSD500 of $\sigma=30$  are listed in Table~\ref{Table: comparisons of different search settings}.

As shown in Table~\ref{Table: comparisons of different search settings}, from \OurMethod-ws to \OurMethod-w40, PSNR and SSIM show slight improvement, 0.01 for PSNR and 0.0003 for SSIM.  Meanwhile the corresponding number of parameters is increased by 52\%. \OurMethod-wm shows the worst performance, and yet it contains the most parameters. With searching for the outer layer width, \OurMethod-ws achieves the best trade-off between the number of parameters and accuracy.

\subsection{Benefits of Using ${\rm {l}_{ssim}}$ Loss}

\newcolumntype{C}[1]{>{\centering\let\newline\\\arraybackslash\hspace{-11pt}}m{#1}}
\begin{table}
\footnotesize
\renewcommand\arraystretch{1.0}
\hspace{-16pt}
\begin{center}
{
\begin{tabular}{l|C{0.7cm}C{0.6cm}C{0.6cm}C{0.6cm}C{0.6cm}C{0.5cm}}
\hline
 \multirow{2}*{Methods}  &   \multicolumn{2}{c}{$\sigma=30$} & \multicolumn{2}{c}{$\sigma=50$} &  \multicolumn{2}{c}{$\sigma=70$}  \\
         &  PSNR  & SSIM   &  PSNR  & SSIM   &  PSNR  & SSIM \\
\hline
N3Net~\cite{plotz2018neural}  & 28.66  & 0.8220  & 26.50  & 0.7490   & 25.18  & 0.6960  \\
\OurMethod\!\!$^*$ & 29.03  & 0.8254  & 26.77  & 0.7498   & 25.42  & 0.6962  \\
\OurMethod\!\!$^{**}$ & \textbf{29.14}  & \textbf{0.8403}  & \textbf{26.77}  & \textbf{0.7635}   & \textbf{25.48}  & \textbf{0.7129}  \\

\hline
\end{tabular}
}
\end{center}
\vspace{-0.1cm}
\caption{Ablation study on BSD500. \OurMethod\!\!$ ^ *$ is trained with single loss MSE and \OurMethod\!\!$ ^{**} $ is trained with the combination loss MSE and ${\rm {l}_{ssim}}$.  }
\vspace{-0.6 cm}
\label{Table: Ablation_study_BSD500}
\end{table}
Here we analyze how our designed loss item ${\rm {l}_{ssim}}$ improves  image restoration results. We implement two baselines: 1) \OurMethod\!\!$^*$ trained with single MSE loss; and 2) \OurMethod\!\!$^{**}$  trained with the combination MSE loss  and ${\rm {l}_{ssim}}$. Table 5 shows the results of these two methods and  that of N3Net on the BSD500 dataset. It is clear that both \OurMethod\!\!$^*$ and  \OurMethod\!\!$^{**}$ outperform the competitive model, while \OurMethod\!\!$^{**}$ trained with the combination loss
shows even better results
over \OurMethod\!\!$^{*}$.

\subsection{Architecture Analysis}

\begin{figure}[t]
\begin{center}
\includegraphics[width=2.75 in]{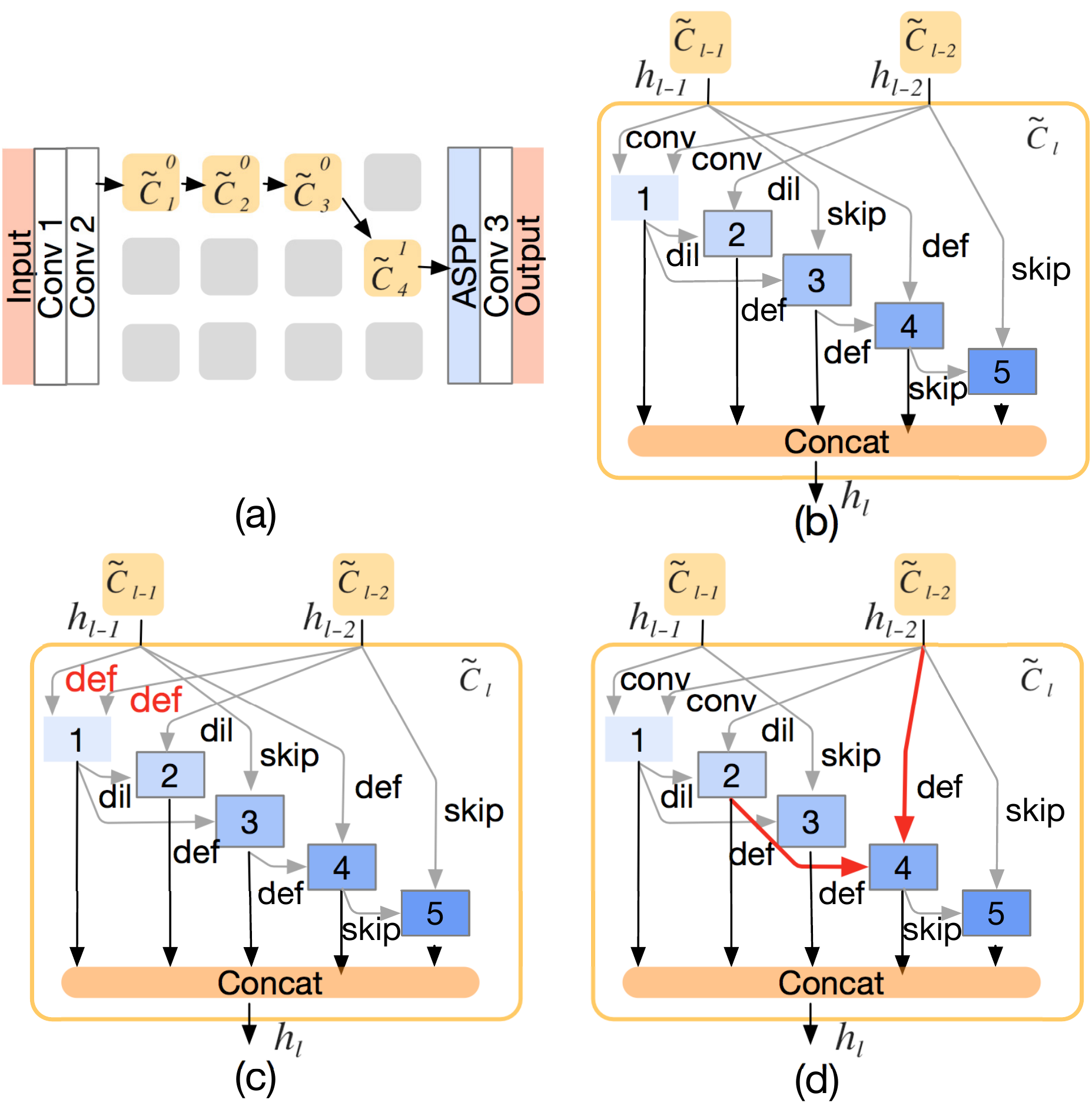}
\end{center}
\vspace{-0.4cm}
\caption{Architecture analysis. `Conv', `def' and `dil' denote conventional, deformable and dilated convolutions. `Skip' is skip connection. (a) Outer layer architecture; (b) inner cell architecture; (c) modified cells, $R1$; (d) modified cells, $R2$.}
 \vspace{-0.2cm}
\label{fig:architecture_analysis}
\end{figure}

Now let us analyse the architectures designed by \OurMethod. Figures~\ref{fig:architecture_analysis} (a) and (b) show the search results in outer network level and the details inside cells, respectively. From Figures~\ref{fig:architecture_analysis} (a) and (b), we can see that:
\begin{enumerate}
\itemsep -0.1cm
    \item In the denoising network found by our \OurMethod, the width of cell that is most close to output layer has the maximum number of channels.
    This  is consist with previous manually designed networks.

    \item Generally speaking, with the same widths, deformable convolution is more flexible and powerful than other convolution operations. Even so, inside cells, instead of connecting all the nodes with the powerful deformable convolution, \OurMethod connects different nodes with different types of operators, such as conventional convolution, dilated convolution and skip connection. We believe that these results prove that \OurMethod is able to select proper operators.

    \item Separable convolutions are not included in the searched results. We conjecture that this is caused by the fact that we do not limit FLOPS or number of parameters  during search. Interestingly, the networks found by our \OurMethod still have fewer parameters than other manual  models.

\end{enumerate}

\begin{table}[t]
\footnotesize
\renewcommand\arraystretch{1.0}
\begin{center}
{
\begin{tabular}{l|ccc}
\hline
Methods &  \OurMethod  & \OurMethod, $R1$ & \OurMethod, $R2$ \\
\hline
PSNR    &  \textbf{29.14}   & 29.06  &  29.13  \\
SSIM    &  \textbf{0.8403}  & 0.8398 &  0.8400 \\
\hline
\end{tabular}
}
\end{center}
\vspace{-0.1cm}
\caption{Architecture analysis.}
\vspace{-0.6 cm}
\label{Table: architecture_analysis}
\end{table}

\begin{table*}[t]
\footnotesize
\renewcommand\arraystretch{1.1}
\begin{center}
{
\begin{tabular}{l|c|cccccc|ccc}
\hline
\multirow{2}*{Methods} & \multirow{2}*{\# parameters (M)} &  \multicolumn{2}{c}{$\sigma$ = 30} & \multicolumn{2}{c}{$\sigma$ = 50} & \multicolumn{2}{c|}{$\sigma$ = 70}  & \multirow{2}*{GPU} & time cost & \multirow{2}*{search method}\\
        &    &  PSNR  & SSIM  & PSNR  & SSIM & PSNR & SSIM &   & (hours)  &   \\
\hline
E-CAE~\cite{suganuma2018exploiting} & 1.05  &  28.23 & 0.8047 & 26.17 & 0.7255 & 24.83  & 0.6636  & 4 Tesla P100  & 96 & EA \\
\OurMethod & 0.63  &  29.14 & 0.8403 & 26.77 & 0.7635 & 25.48  & 0.7129  & 1 Tesla V100  & 16.5 & gradient \\
\hline
\end{tabular}
}
\end{center}
\vspace{-0.2cm}
\caption{Comparisons with E-CAE on BSD500.}
\vspace{-0.4 cm}
\label{Table: comparison with E-CAE}
\end{table*}

\begin{table*}[h]
\footnotesize
\renewcommand\arraystretch{1.0}
\begin{center}
{
\begin{tabular}{l|c|c|cc|cc|cc}
\hline
\multirow{2}*{Methods}  & \multirow{2}*{\# parameters (M)}  & \multirow{2}*{time cost (s)} & \multicolumn{2}{c|}{$\sigma=30$} & \multicolumn{2}{c|}{$\sigma=50$} & \multicolumn{2}{c}{$\sigma=70$} \\
                             &      &     & PSNR   & SSIM    & PSNR   & SSIM    & PSNR   &  SSIM    \\
\hline
BM3D~\cite{dabov2007image}   &   -  &  -  & 27.31  & 0.7755  & 25.06  & 0.6831  & 23.82  &  0.6240  \\
WNNM~\cite{gu2014weighted}   &   -  &  -  & 27.48  & 0.7807	 & 25.26  &	0.6928	& 23.95  &  0.3460  \\
RED~\cite{mao2016image}      & 0.99 &  -  & 27.95  & 0.8056  & 25.75  &	0.7167	& 24.37  &  0.6551  \\
MemNet~\cite{tai2017memnet}  & 4.32 &  -  & 28.04  & 0.8053	 & 25.86  &	0.7202	& 24.53  &  0.6608  \\
NLRN~\cite{liu2018non}       & 0.98 &  10411.49  & 28.15  & \textbf{0.8423} & 25.93  & 0.7214	& 24.58  &  0.6614  \\
E-CAE~\cite{suganuma2018exploiting} & 1.05 & - & 28.23 & 0.8047  & 26.17  & 0.7255  & 24.83  & 0.6636 \\
DuRN-P~\cite{liu2019dual}    & 0.78 &  -  & 28.50  & 0.8156  & 26.36  & 0.7350  & 25.05  &  0.6755\\
N3Net~\cite{plotz2018neural} & 0.68 &  121.11  & 28.66  & 0.8220  & 26.50  & 0.7490	& 25.18  &  0.6960  \\
\OurMethod                     & \textbf{0.63} &  \textbf{83.25}  & \textbf{29.14}  & 0.8403  & \textbf{26.77}  &	\textbf{0.7635}  & \textbf{25.48}  & \textbf{0.7129} \\
\hline
\end{tabular}
}
\end{center}
\vspace{-0.2cm}
\caption{Denoising experiments. Comparisons with state-of-the-arts on the BSD500 dataset. We show our results in the last row. Time cost means GPU-seconds for inference on the 200 images from the test set of BSD500 using one single GTX 980 graphic card.}
\vspace{-0.8 cm}
\label{Table: BSD500}
\end{table*}

From Figure~\ref{fig:architecture_analysis} (b), we can see that the networks found  by \OurMethod consist of many fragmented branches, which might be the main reason why the designed networks have better performance than previous denoising models. As explained in~\cite{ma2018shufflenet}, the fragmentation structure is beneficial for accuracy. Here we verify if \OurMethod improves the accuracy by designing a proper architecture or by simply integrating various branch structures and convolution operations. We modify the architecture found by our \OurMethod in two different ways and then compare the modified architectures with unmodified architectures.

The first modification is replacing conventional convolutions in the searched architectures with deformable convolutions as shown in Figure~\ref{fig:architecture_analysis} (c). As  mentioned above, deformable convolution is more flexible than conventional convolution, replacing conventional convolutions with deformable convolutions \textit{in theory} should improve the capacity of networks. The other modification is to change the connection relationships between nodes inside each cell, as shown in Figure~\ref{fig:architecture_analysis} (d), which is aiming to verify if the connection relationship built by our \OurMethod is indeed appropriate.

The modification parts are marked in red in Figure~\ref{fig:architecture_analysis} (c) and (d). Following the two proposed modifications, we also modify other parts for comparison experiments. However, limited by space, we only show two examples  here. The comparison results are listed in Table~\ref{Table: architecture_analysis}, where the two mentioned modification operations, are denoted as $R1$ and $R2$.

From Table~\ref{Table: architecture_analysis}, we can see that both
modifications  reduce the accuracy. Replacing convolution operation reduces the PSNR and SSIM by 0.08 and 0.0005, respectively. Changing connection relationships decreases the PSNR and SSIM to 29.13 and 0.8400, respectively.

From the comparison results, we can draw a conclusion:  \OurMethod does find  a proper structure and select proper convolution operations, instead of simply integrating a complex network with various operations.
\textit{The fact that a slight  perturbation to the found architecture deteriorates the accuracy
indicates that the found architecture is indeed a local optimum in the architecture search space.
}

\subsection{Comparisons with Other NAS Methods}
\label{Sec: comparisons with NAS methods}

Inspired by recent advances in NAS, three NAS methods have been proposed for low-level image restoration tasks ~\cite{suganuma2018exploiting, chu2019fast, liu2019deep}. E-CAE~\cite{suganuma2018exploiting} is proposed for image inpainting and denoising. FALSR~\cite{chu2019fast} is proposed for super resolution. EvoNet~\cite{liu2019deep} searches for networks for medical image denoising. All three methods are based on EA and require a large amount of compuational resources and %
are GPU time hungry.
By using four P100 GPUs, E-CAE takes four days (384 GPU hours) to execute the evolutionary algorithm and fine-tune the best model for denoising on BSD500. FALSR takes about 3 days on 8 Tesla-V100 GPUs (576 GPU hours) to find the best architecture. EvoNet uses 4 Geforce TITAN GPUs and takes 135 hours for finding the best gene.
Here we mainly focus on comparing our \OurMethod with E-CAE, because both them are proposed for searching for architectures for the task of denoising on BSD500.
Table~\ref{Table: comparison with E-CAE} shows the details.

Compared with E-CAE~\cite{suganuma2018exploiting}, FALSR~\cite{chu2019fast} and EvoNet~\cite{liu2019deep}, our \OurMethod is much faster in searching. By using a single Tesla V100, \OurMethod takes about 4.5 hours in searching and 12 hours for training the network found by our algorithm. The fast search speed of our \OurMethod benefits from the following three advantages.
\begin{enumerate}
\itemsep -0.1cm

    \item \OurMethod uses a gradient based search strategy. EA based NAS methods generally need to train a large number of children networks (genes) to update their populations. For instance, FALSR trained about 10k models during its searching process.
    In sharp contrast,
    our \OurMethod only needs to train one supernet in the search stage.

    \item In searching for the outer layer width, we share cells across different feature levels, saving  memory consumption in the supernet. As a result, we can use
    larger
    batch sizes for  training the supernet, which further speeds up search.

    \item By using a simple early-stopping search strategy, \OurMethod further saves 0.5 to 1.5 hours in the search stage.

\end{enumerate}

\subsection{Comparisons with State-of-the-art}

\begin{figure}[b!]
\begin{center}
\includegraphics[width=3.1 in]{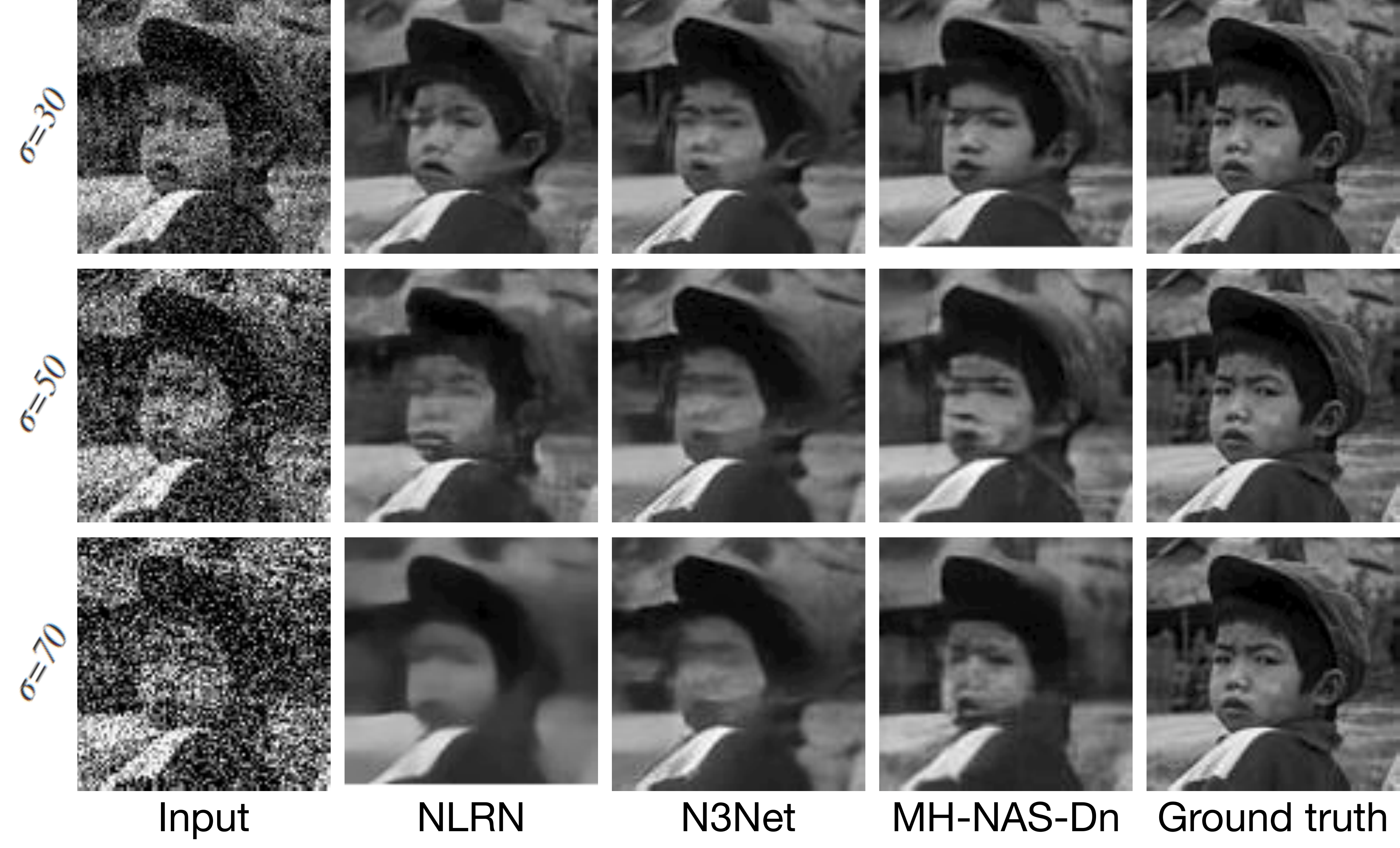}
\end{center}
\vspace{-0.4cm}
\caption{Denoising experiments on  BSD500.}
\vspace{-0.2cm}
\label{fig:comparison_BSD500}
\end{figure}

\begin{figure}[h]
\begin{center}
\includegraphics[width=3.1 in]{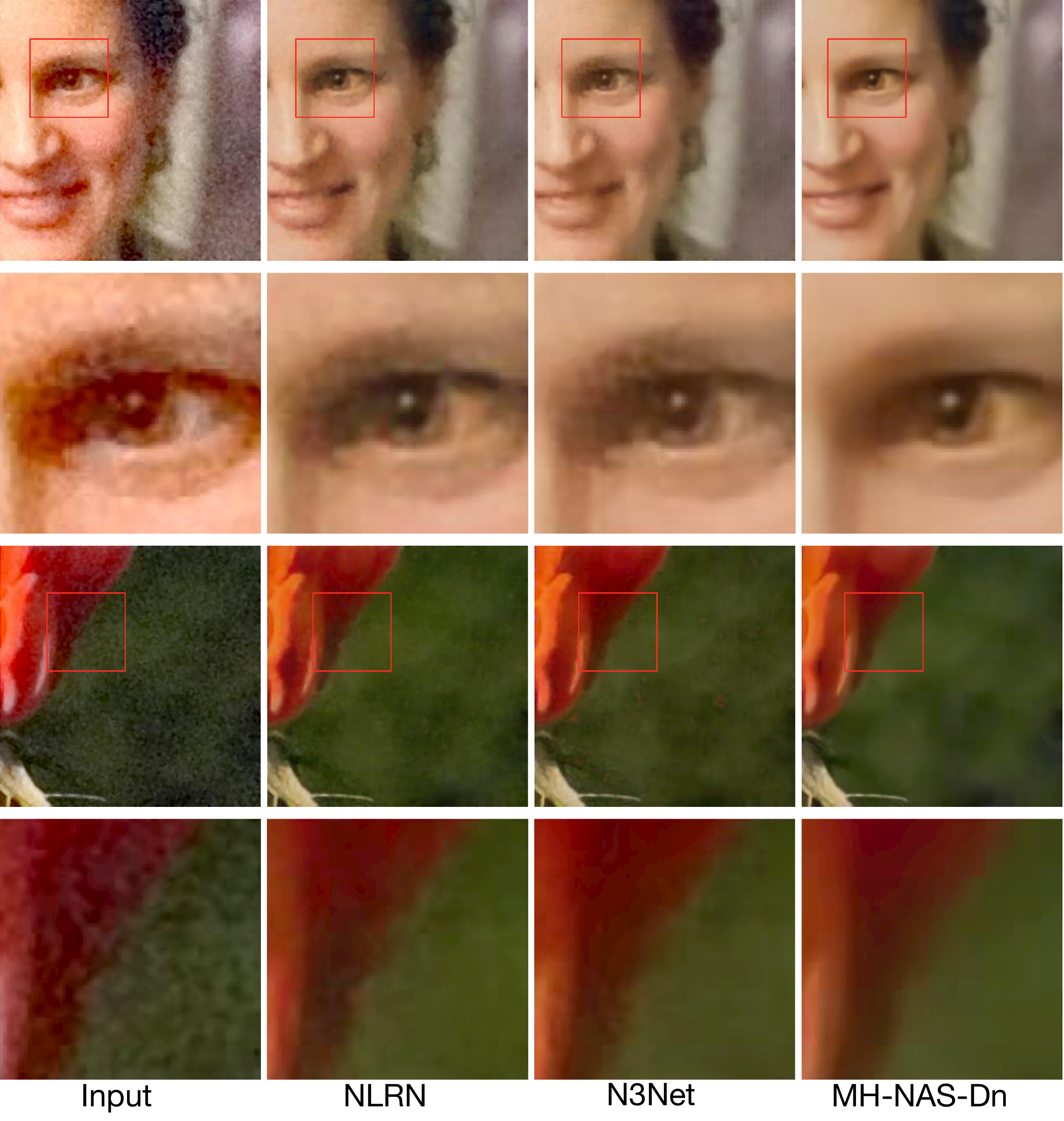}
\end{center}
\vspace{-0.4cm}
\caption{Denoising experiments on SIM1800.}
\vspace{-0.2cm}
\label{fig:comparison_sim}
\end{figure}

\begin{table}
\footnotesize
\renewcommand\arraystretch{1.0}
\begin{center}
{
\begin{tabular}{l|cc}
\hline
Methods                          &  PSNR   & SSIM   \\
\hline
NLRN~\cite{liu2018non}           &  27.53  & 0.8081 \\
N3Net~\cite{plotz2018neural}     &  \textbf{27.62} & 0.8191  \\
\OurMethod                           &  27.23  & \textbf{0.8326} \\
\hline
\end{tabular}
}
\end{center}
\vspace{-0.2cm}
\caption{Denoising results on SIM1800.}
\vspace{-0.7 cm}
\label{Table: SIM1800}
\end{table}

Now we compare the \OurMethod designed networks with a number of recent methods and use PSNR and SSIM to quantitatively measure the restoration performance of those methods. The comparison results on BSD500 and SIM1800 are listed in Table~\ref{Table: BSD500} and Table~\ref{Table: SIM1800}, respectively. Figures~\ref{fig:comparison_BSD500} and~\ref{fig:comparison_sim} show the visualization.

Table~\ref{Table: BSD500} shows that N3Net and \OurMethod beat other models by a clear margin. Our proposed \OurMethod achieves the best performance when $\sigma$ is set to 50 and 70. When the noise level $\sigma$ is set to 30, the SSIM of NLRN is slightly higher (0.002) than that of our \OurMethod, but the PSNR of NLRN is much lower (nearly 1dB) than that of \OurMethod.

\textit{Overall our \OurMethod achieves better performance than others. In addition, compared with the second best model N3Net, the network designed by \OurMethod has fewer parameters and is faster in inference.} As listed in Table~\ref{Table: BSD500}, the \OurMethod designed network has 0.63M parameters, which is 92.65\% that of N3Net and 60\% that of E-CAE. Compared with N3Net, the \OurMethod designed network reduces the inference time on the test set of BSD500 by \textbf{31.26\%}.
We compare the network designed by \OurMethod with NLRN and N3Net on SIM1800. Table~\ref{Table: SIM1800} lists the results, from which we can see that the SSIM of the \OurMethod desgined network is much higher than that of NLRN and N3Net. However, PSNR of the \OurMethod designed network is slightly lower than that of NLRN and N3Net. In summary, the performance of the \OurMethod designed network is competitive with that of NLRN and N3Net on SIM1800. Figure~\ref{fig:comparison_sim} shows a visual comparison.

\begin{figure}
\begin{center}
\includegraphics[width=3.10 in]{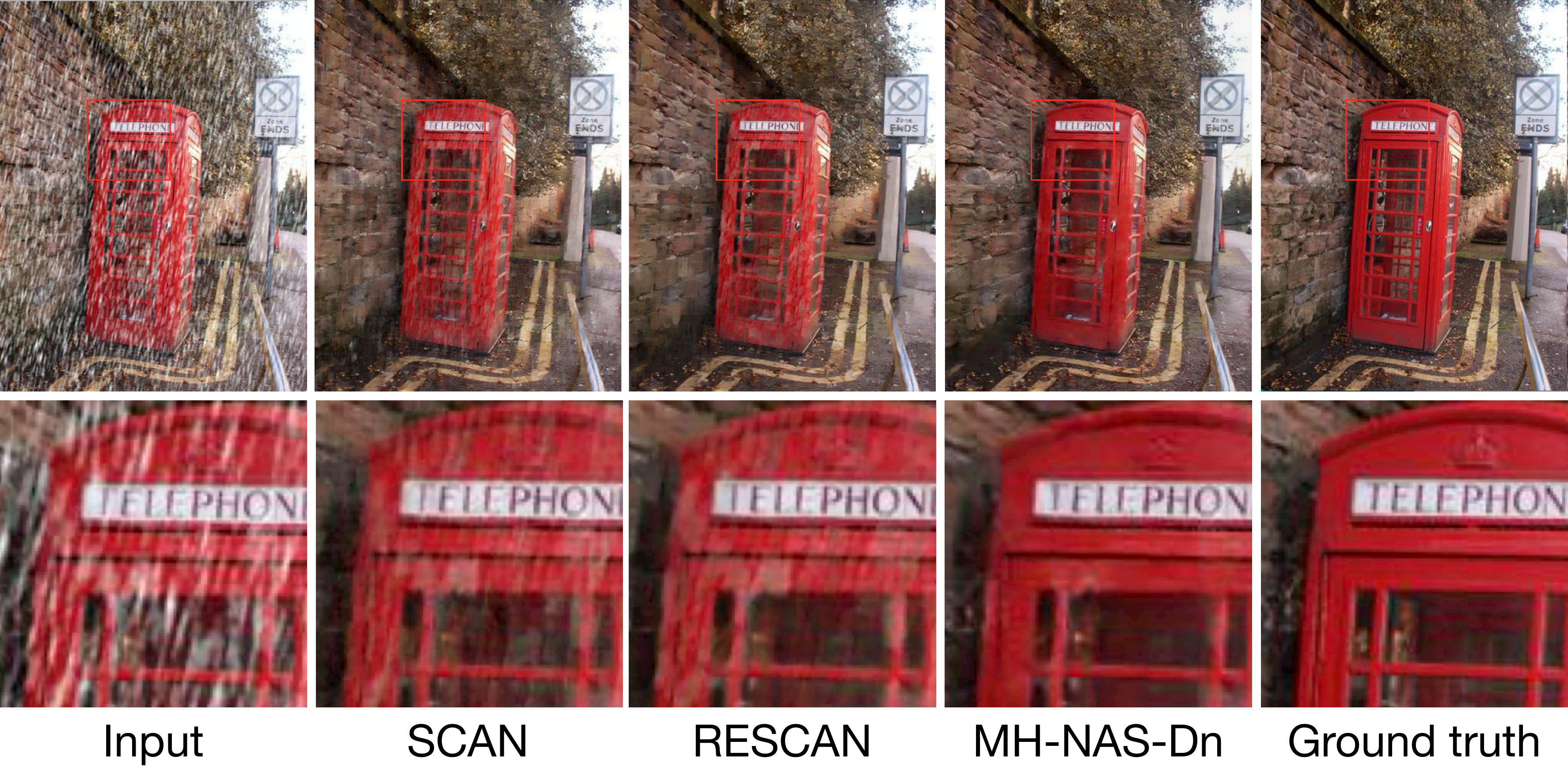}
\end{center}
\vspace{-0.4cm}
\caption{De-raining experiments on Rain800.}
\vspace{-0.2cm}
\label{fig:comparison_rain800}
\end{figure}

\begin{table}[t]
\footnotesize
\renewcommand\arraystretch{1.0}
\begin{center}
{
\begin{tabular}{l|cc}
\hline
Methods                          &  PSNR   & SSIM   \\
\hline
DSC~\cite{luo2015removing}       &  18.56  & 0.5996 \\
LP~\cite{li2016rain}             &  20.46  & 0.7297 \\
DetailsNet~\cite{fu2017removing} &  21.16  & 0.7320 \\
JORDER~\cite{yang2017deep}       &  22.24  & 0.7763 \\
JORDER-R~\cite{yang2017deep}       &  22.29  & 0.7922 \\
SCAN~\cite{li2018recurrent}      &  23.45  & 0.8112 \\
RESCAN~\cite{li2018recurrent}    &  24.09  & 0.8410 \\
\OurMethod                          &  \textbf{26.31}  & \textbf{0.8685} \\
\hline
\end{tabular}
}
\end{center}
\vspace{-0.2cm}
\caption{De-raining results on Rain800. With a GTX 980 graphic card, RESCAN and \textbf{\OurMethod} respectively cost 44.35, \textbf{21.80} GPU-seconds for inference on the test set of Rain800.}
\vspace{-0.7 cm}
\label{Table: Rain800}
\end{table}

\noindent
\textbf{Additional experiments}
We apply the proposed \OurMethod on a challenging de-raining dataset Rain800. The supernet that we build for image de-raining contains 3 cells and each cell is made up of 4 nodes. Search and training setting are consistent with that of the denoising experiments, except that we use random crop and horizontal flipping for augmentation. The results are listed in Table 3 and shown in Figure 5. As shown in Table 3, the de-raining network designed by \OurMethod achieves  much better performance than others. Comparing RESCAN to the network designed by \OurMethod, PSNR and SSIM are improved by 2.22 and 0.0275, respectively. In addition, the inference speed of \OurMethod designed de-raining network is 2.03$\times$ that of RESCAN.

\section{Conclusion}

In this work, we have proposed \OurMethod, an memory-efficient hierarchical architecture search algorithm for the low-level image restoration task image denoising. \OurMethod adopts differentiable architecture search algorithms and a cell sharing strategy. It is both memory and computation efficient, taking only about 4.5 hours to search using a single GPU.
In addition, a simple but effictive early stopping strategy is used to avoid the NAS collapse problem. Our proposed \OurMethod achieves highly competitive or better performance compared with previous state-of-the-art methods with fewer parameters and a faster inference speed.
We believe that the proposed method can be applied to many other low-level image processing tasks.

\textbf{Acknowledgments}
C. Shen's participation was in part supported by the ARC Grant ``Deep learning that scales''.

{\small
\bibliographystyle{ieee_fullname}
\bibliography{egbib}
}

\end{document}